\documentclass[runningheads]{llncs}
\usepackage{graphicx}
\usepackage{hyperref}
\usepackage{url}
\usepackage{listings}
\usepackage{rotating}
\usepackage{multirow}

\usepackage{algorithm}
\usepackage{algorithmic}
\usepackage{amsmath}
\usepackage{amssymb}
\usepackage{mathtools}
\usepackage{subcaption}
\usepackage{multirow}
\usepackage{cite}
\usepackage{xcolor}
\begin{document}
\title{MixQuant: Mixed Precision Quantization with a Bit-width Optimization Search}
\titlerunning{MixQuant}
%
\author{Eliska Kloberdanz\orcidID{0000-0001-7159-2937} \and
Wei Le}
\authorrunning{E. Kloberdanz et al.}
%
\institute{Department of Computer Science, Iowa State University, Ames IA 50011, USA \\
\email{\{eklober,weile\}@iastate.edu}}
\maketitle              
\begin{abstract}
Quantization is a technique for creating efficient Deep Neural Networks (DNNs), which involves performing computations and storing tensors at lower bit-widths than f32 floating point precision. Quantization reduces model size and inference latency, and therefore allows for DNNs to be deployed on platforms with constrained computational resources and real-time systems. However, quantization can lead to numerical instability caused by roundoff error which leads to inaccurate computations and therefore, a decrease in quantized model accuracy. Similarly to prior works, which have shown that both biases and activations are more sensitive to quantization and are best kept in full precision or quantized with higher bit-widths, we show that some weights are more sensitive than others which should be reflected on their quantization bit-width. To that end we propose \textit{MixQuant}, a search algorithm that finds the optimal custom quantization bit-width for each layer weight based on roundoff error and can be combined with any quantization method as a form of pre-processing optimization. We show that combining \textit{MixQuant} with BRECQ, a state-of-the-art quantization method, yields better quantized model accuracy than BRECQ alone. Additionally, we combine \textit{MixQuant} with vanilla asymmetric quantization to show that \textit{MixQuant} has the potential to optimize the performance of any quantization technique.

\end{abstract}
\section{Introduction}
Quantization is a method for mapping continuous values to a set of discrete values. The goal of neural network quantization is to perform computations and store tensors at lower bit-widths than floating point precision to reduce model size and inference latency while maintaining model accuracy, which allows for deploying DNNs on platforms with constrained computational resources, e.g.: real time inference on mobile devices. Quantization can be performed during training or inference. In this paper we focus on quantized inference, specifically post-training quantization, which quantizes a full precision trained model without the need for re-training or fine-tuning.

Quantized inference can be either simulated or integer-only, and in this paper we focus on simulated quantization, where the quantized model parameters are stored in low-precision, but the mathematical operations on them (e.g. matrix multiplications and additions) are performed with floating point arithmetic \cite{Gholami2022ASO}. In Tensorflow, PyTorch, and HuggingFace (QDQBERT model), simulated quantization is referred to as fake quantization. This means that the DNN parameters are first quantized from f32 to, for example, int4, and then dequantized back to f32 to perform the forward pass executed during inference. We show that the roundtrip process of quantizing and dequantizing the model parameters leads to roundoff error, which may lead to numerical instability.

Similarly to prior works, which have shown that both biases and activations are more sensitive to quantization and are best kept in full precision or quantized with higher bit-widths \cite{Zhou2016DoReFaNetTL}, we show that some weights are more sensitive than others which should be reflected on their quantization bit-width. To that end we propose \textit{MixQuant}, a search algorithm that finds the optimal quantization bit-width from int2, int3, int4, int5, int6, int7, and int8 for each layer weight based on roundoff error and can be combined with any quantization method as a form of pre-processing optimization. We show that combining \textit{MixQuant} with BRECQ \cite{Li2021BRECQPT}, a state-of-the-art quantization method, yields better quantized model accuracy than BRECQ alone. Additionally, we combine \textit{MixQuant} with vanilla asymmetric quantization to show that \textit{MixQuant} has the potential to optimize the performance of any quantization technique.

\textit{MixQuant} has three main benefits. First, \textit{MixQuant} is a component of the quantization process, which can be leveraged to find optimal quantization mixed precision bit-widths that can be plugged into any quantization method to optimize its performance. Second, \textit{MixQuant} is linear and runs in a matter of seconds, which makes it practical. Third, combining \textit{MixQuant} with BRECQ, a state-of-the-art quantization method yields better quantized model accuracy than BRECQ alone, OMSE \cite{Choukroun2019LowbitQO}, AdaRound \cite{Nagel2020UpOD}, AdaQuant \cite{Hubara2020ImprovingPT}, and Bit-Split \cite{Wang2020TowardsAP}.

\section{Related Work}
\subsection{Neural Network Quantization} 
Neural network quantization can be applied to training \cite{Gupta2015DeepLW,  Zhou2016DoReFaNetTL, Hubara2017QuantizedNN, Bartan2021TrainingQN, Elthakeb2020DivideAC} or inference. There are two paradigms in quantized DNN inference: post-training quantization (PTQ) and quantization-aware training (QAT) \cite{Jacob2018QuantizationAT, Tailor2021DegreeQuantQT}. In contrast to PTQ, QAT requires that the f32 model is retrained while simulating quantized inference in the forward pass. While \textit{MixQuant} can be integrated with either, we focus on PTQ which does not require any re-training.

\cite{Hubara2021AccuratePT} and \cite{Li2021BRECQPT} are amongst the recent state-of-the-art post training quantization works.  \cite{Hubara2021AccuratePT} introduce AdaQuant, which finds optimal quantization for both weights and activations and is based on minimizing the error between quantized layer outputs and f32 layer outputs. This approach is similar to \textit{MixQuant}; however, \textit{MixQuant} finds the optimal quantization bit-widths based on quantization error (QE) minimization, while AdaQuant treats the bit-width as a constant and quantizes all weights and activations using the same bit-width (either \textit{int8} or \textit{int4}). \cite{Li2021BRECQPT} propose BRECQ, a quantization method based on DNN block reconstruction. \cite{Nagel2020UpOD} propose AdaRound, adaptive rounding for weights, which achieves better accuracy than rounding to the nearest. They formulate the rounding procedure as an optimization problem that minimizes the expected difference between model loss with and without weights quantization perturbation. \cite{Li2020AdditivePQ} develop a method based on constraining all quantization levels as the sum of Powers-of-Two terms, \cite{Wang2020TowardsAP} propose a Bit-Split and Stitching framework (Bit-split), \cite{Nahshan2021LossAP} study the effect of quantization on the structure of the loss landscape, \cite{Banner2019PostT4} develop ACIQ-Mix, a 4 bit convolutional neural network quantization, and \cite{Cai2020ZeroQAN} perform zero-shot quantization ZeroQ based on distilling a dataset that matches the input data distribution. 

Quantization originated with convolutional neural networks, but it has been extended to natural language processing neural networks as well. \cite{Chen2020DifferentiablePQ} propose differentiable product quantization, a learnable compression for embedding layers in DNNs. \cite{Kim2021IBERTIB} study an integer-only quantization scheme for transformers, where the entire inference is performed with pure integer arithmetic.

Other works studied hardware optimization for quantization or the relationship between quantization and adversarial robustness. \cite{Han2020ExtremelyLC} focus on performance optimization for Low-bit Convolution on ARM CPU and NVIDIA GPU. \cite{Fu2021DoubleWinQA} investigate quantized models' adversarial robustness. They find that when an adversarially trained model is quantized to different precisions in a post-training manner, the associated adversarial attacks transfer poorly between different precisions. 

\subsection{Mixed Precision Quantization} 
In this paper we focus on mixed precision quantization. There are only a few prior works that focus on mixed precision quantization since most focus on single precision quantization, where the quantization bit-width of all weights are uniform and therefore; treated as a constant. \cite{Wang2019HAQHA} propose a framework for determining the quantization policy with mixed precision and reinforcement learning, but compared to \textit{MixQuant} it requires significantly more overhead (hardware simulators and reinforcement learning). \cite{Liang2020PostTM} focuses on mixed precision quantization of activations and distinguishes between key and non-key activations to assign 8-bit and 4-bit precision respectively. In contrast to MixQuant, which searches for weights mixed precision from 8 to 2 bits, \cite{Liang2020PostTM} is limited to a choice between 4 and 8 bits and applies only to activations while all weights are quantized with 8-bit precision.  The primary focus of \cite{Wu2018MixedPQ} is neural architecture search, which can also be used for mixed precision quantization. However, their search on ResNet 18 for ImageNet takes 5 hours, while \textit{MixQuant} runs in order of a few seconds. \cite{Liu2021PosttrainingQW} use single precision for weights, where the mixed precision is represented only by selecting a different bit-width for weights than activations. \cite{Liu2021PosttrainingQW} is the most most recent, and we show that \textit{MixQuant} yields better accuracy.

Another mixed precision quantization work that we build on is \cite{Lin2016FixedPQ}, who identify optimal bit-width allocation across DNN layers. However, there are two primary differences between \cite{Lin2016FixedPQ} and our work: (1) \cite{Lin2016FixedPQ} focus on fixed-point precision, not integer precision, (2) \cite{Lin2016FixedPQ} a different method for finding layer bit-widths based on predicted signal-quantization-to-noise -ratio. Moreover, while they find that on CIFAR-10 convolutional DNN is able to achieve 20 \% model size reduction; their AlexNet experiments on ImageNet-1000 achieve less than 1\% model reduction. In this work we are able to successfully leverage mixed precision optimal bit-width allocation on ImageNet-1000 models.

\section{Quantization and Numerical Instability}
Quantization involves lowering the bit-width of a numeric tensor representation, which can cause numerical instability that leads to inaccurate outputs \cite{Kloberdanz2022DeepStabilityAS}. In general, numerical instability arises due to two types of numerical errors: (1) roundoff errors and (2) truncation errors. Roundoff errors are caused by approximating real numbers with finite precision, while truncation errors are caused by approximating an a mathematical process such as the Taylor series. We argue that quantization can significantly amplify the roundoff error, which leads to a degradation in quantized DNN accuracy.

DNN training and inference is typically performed in f32 precison, which already introduces roundoff errors, because it has only 32 bits to represent real numbers. Specifically, f32 can represent a zero and numbers from -3.40282347E+38 to -1.17549435E-38 and from 1.17549435E-38 to 3.40282347E+38, but numbers outside of this range are not representable in f32. In simulated quantization the process of quantizing DNN parameters from f32 to int (e.g.: int4) and dequantizing them back to f32 to perform matrix multiply and add (e.g.: inputs * weights + biases) can lead to a loss of precision. 

Listing 1 shows an example of a simple simulated quantized inference, where the weights tensor is quantized to int2 and its subsequent dequantization back to f32 has a roundoff error. The roundoff error occurs in the second element of the weight tensor, which becomes 0.0 (line 40) while its true original value is 0.01 (line 38). This error caused by quantization then propagates further - the computation $inputs * weights + biases$ returns 1.0000e-05 (line 43) instead of 1.2000e-05 (line 42) in the second element of the result tensor.

\lstset{escapeinside={<@}{@>}}
\begin{lstlisting}[basicstyle = \tiny, frameround=\thinlines, xleftmargin=.06\textwidth, numbers=left, caption = Loss of Precision due to Quantization Example]
def scale(r, bits):
    min_r = r.min()
    max_r = r.max()
    qmin = -1 * (2 ** (bits - 1))
    qmax = 2 ** (bits - 1) - 1
    scale_r = (max_r - min_r) / (qmax - qmin)
    return scale_r

def zero_point(r, bits):
    scale_r = scale(r, bits)
    min_r = r.min()
    qmin = -1 * (2 ** (bits - 1))
    zpt_r = qmin - int(min_r / scale_r)
    return zpt_r

def quant(r, bits):
    z = zero_point(r, bits)
    s = scale(r, bits)
    q = (torch.round(r/s) + z).int()
    return q

def dequant(q, z, s):
    r = (q - z) * s
    return r.float()

input = torch.tensor([0.005, 0.0002, 0.01, 0.003])
bias = torch.tensor([0.00001])
weight = torch.tensor([-1.0, 0.01, 1.0, 2.0]) # original weight

S = scale(weight, 2) # quantization scale
Z = zero_point(weight, 2) # quantization zero point
q_weight = quant(weight, 2) # quantized weight
dq_weight = dequant(q, Z, S) # dequantized weight

result = input * weight + bias
dq_result = input * dq_weight + bias

<@\textbf{f32 weight:}@>  tensor([-1.0000,  <@\textcolor{green}{0.0100}@>,  1.0000,  2.0000])
quantized weight:  tensor([-2, -1,  0,  1], dtype=torch.int32)
<@\textbf{dequantized weight:}@>  tensor([-1.,  <@\textcolor{red}{0.}@>,  1.,  2.])

<@\textbf{f32 result:}@>  tensor([-4.9900e-03,  <@\textcolor{green}{1.2000e-05}@>,  1.0010e-02,  6.0100e-03])
<@\textbf{simulated quantization result:}@>  tensor([-4.9900e-03,  <@\textcolor{red}{1.0000e-05}@>,  1.0010e-02,  6.0100e-03])
\end{lstlisting}

\section{MixQuant}
\textit{MixQuant} is a quantization scheme that relies on mixed precision to find the bit-widths of individual layer weights that minimize roundoff error and therefore, minimize model accuracy degradation due to quantization. Specifically, \textit{MixQuant} is a search algorithm that finds optimal bit-widths that minimize model accuracy degradation caused by quantization. Prior works have shown that biases and activations are more sensitive to quantization than weights, and are therefore typically kept in higher precision. In this paper we argue some weights are more sensitive to quantization than others, which we show in our ablation studies. This warrants a careful bit-width allocation to individual weights and serves as motivation for MixQuant. In essence, \textit{MixQuant} can be viewed as an additional pre-processing optimization component of the quantization process, which can be combined with any quantization method optimize its performance. 

\textit{MixQuant} is described in Algorithm~\ref{alg:MixQuant}. The optimal weight layer bit-widths search has two primary components: layer-wise QE minimization and a QE multiplier (QEM). The layer-wise QE is calculated as the mean squared error (MSE) between the f32 model weights and the weights that have been dequantized following an int quantization (any quantization method can be used at line 8 in Algorithm~\ref{alg:MixQuant}) to capture the information loss due to roundoff error caused by quantization. This error is calculated for each layer for each bit-width from the following list: 8, 7, 6, 5, 4, 3, and 2 (lines 4-11 in Algorithm~\ref{alg:MixQuant}). Following that, \textit{MixQuant} searches for the optimal bit-width for each layer by comparing the QE of each bit-width from this list with an int8 error, which serves as a baseline (lines 12-13 in Algorithm~\ref{alg:MixQuant}). To push \textit{MixQuant} to select bit-widths lower than int8, \textit{MixQuant} leverages the QEM. If the QE at a bit-width {\tt b} is less or equal to int8 QE multiplied by the QEM, {\tt b} becomes the optimal bit-width for that layer. This can be expressed as an optimization problem:
\begin{equation}
  \begin{split}
        optBit &= {\arg\min}_{optBit}\ quantErrors \\           
\text{Subject to}
        &\ quantErrors \leq 8bit_qError * QEM  \\
        &  optBit \in B
    \end{split}  
\end{equation}

Because the QEM is an input parameter into MixQuant, it allows the user to specify a custom trade-off between quantization bit-width and model accuracy; and therefore, it allows the user to find \textit{their} optimal layer bit-width.
\vspace{-5mm}
\begin{algorithm}
   \caption{MixQuant}
   \label{alg:MixQuant}
\begin{algorithmic}[1]
\renewcommand{\algorithmiccomment}[1]{#1}
   \STATE {\bfseries Input:} full precision weights $W$, bit-widths $B$, QE multiplier $QEM$
   \STATE Initialize $optimalBitWidths$ \\
   \COMMENT /* Iterate over all layers */
   \FOR{$l$ {\bfseries in} $layers$}
   \STATE $8bit_W$ = Quantize($W$, $bitWidth=8$) \\
   \COMMENT /* Compute int8 quantization error in layer l */
   \STATE $8bit_qError$ = $W$ - Dequantize($8bit_W$) \\
   \COMMENT /* For every bit-width in $B$ compute quantization error in layer l */
   \STATE Initialize $quantErrors$ 
   \FOR{$bitWidth$ {\bfseries in} $B$}
   \STATE $quantizedW$ = Quantize($W$, $bitWidth$)
   \STATE $qError$ = $W$ - Dequantize($quantizedW$)
   \STATE Append $qError$ to $quantErrors$
   \ENDFOR \\
   \COMMENT/* Select optimal bit-width at layer l */
   \STATE $optBit$ = ${\arg\min}_{optBit}$\ {quantErrors \ s.t.\ \par \setlength{\parindent}{9ex} quantErrors $\leq$ $8bit_qError$ * $QEM$, \par $optBit$ $\in$ $B$}
   \STATE Append $optBit$ to $optimalBitWidths$
   \ENDFOR
   \STATE \textbf{return} $optimalBitWidths$
\end{algorithmic}
\end{algorithm}
\vspace{-10mm}

\paragraph{Weights Mixed Precision Quantization} We focus on weights quantization for three reasons. First, weights account for majority of parameters in a DNN and therefore, have the greatest impact on model size and inference time. Second, model accuracy is more sensitive to quantized activations than weights \cite{Zhou2016DoReFaNetTL}. Third, we guided our algorithm design with the state-of-the art results in \cite{Li2021BRECQPT}, who introduced BRECQ which shows weight-only quantization.

\begin{figure}[]
     \centering
     \captionsetup[subfigure]{justification=centering}
     \begin{subfigure}[b]{0.29\textwidth}
         \centering
         \includegraphics[width=\linewidth]{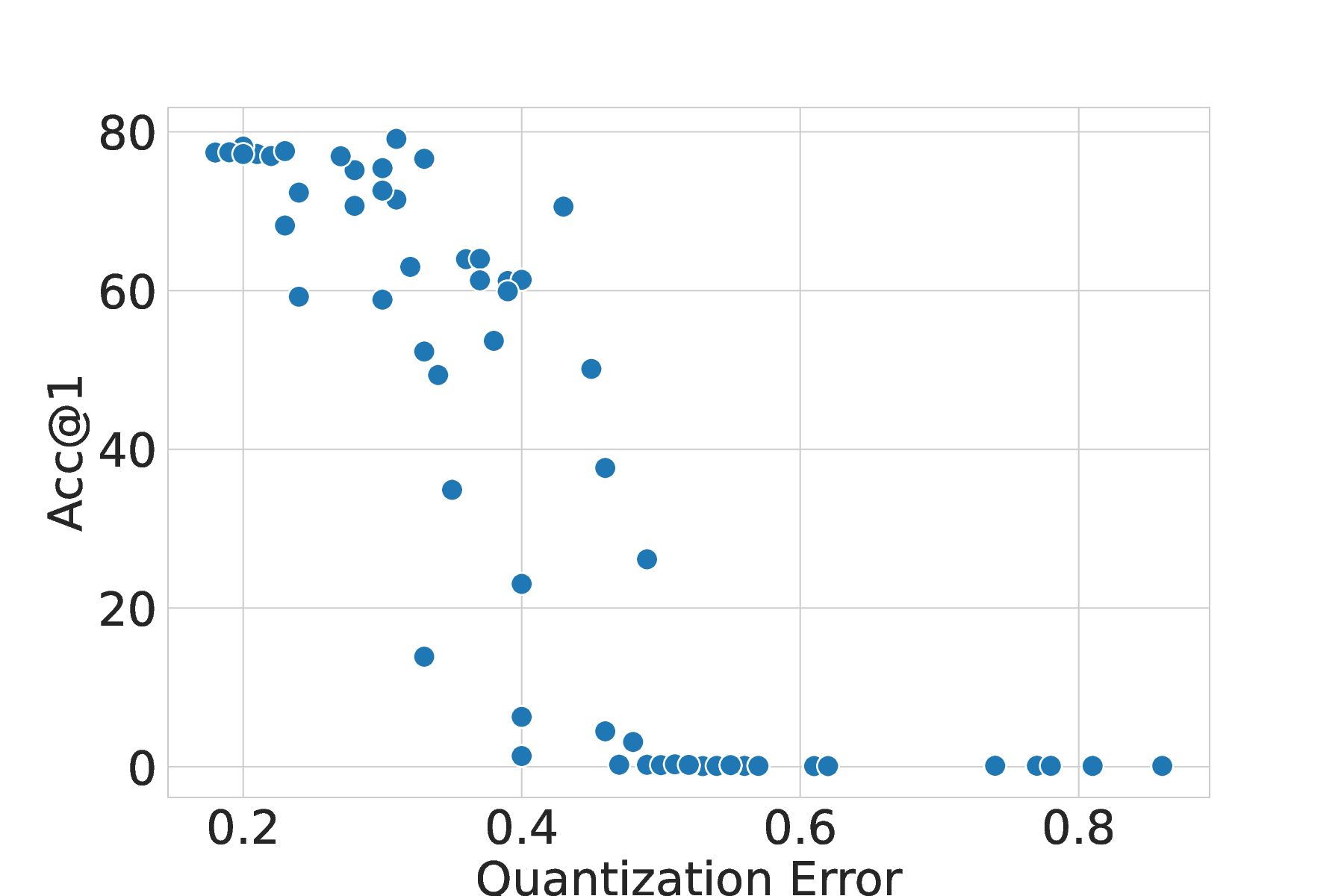}
         \caption{Top 1 accuracy vs. quantization error}
         \label{fig:top1}
     \end{subfigure}
     \hfill
     \begin{subfigure}[b]{0.29\textwidth}
         \centering
         \includegraphics[width=\linewidth]{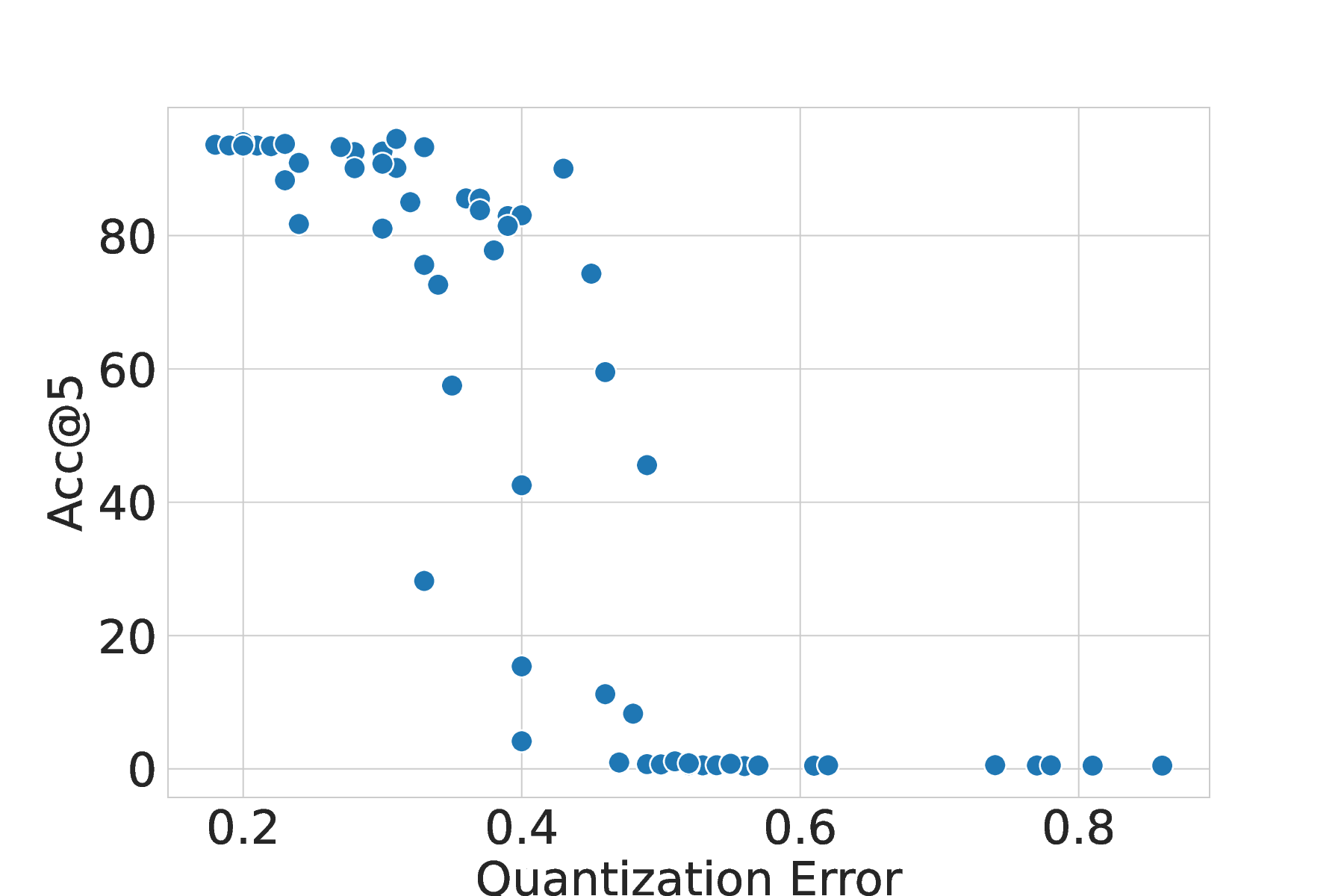}
         \caption{Top 5 accuracy vs. quantization error}
         \label{fig:top5}
     \end{subfigure}
     \hfill
     \begin{subfigure}[b]{0.3\textwidth}
         \centering
         \includegraphics[width=\linewidth]{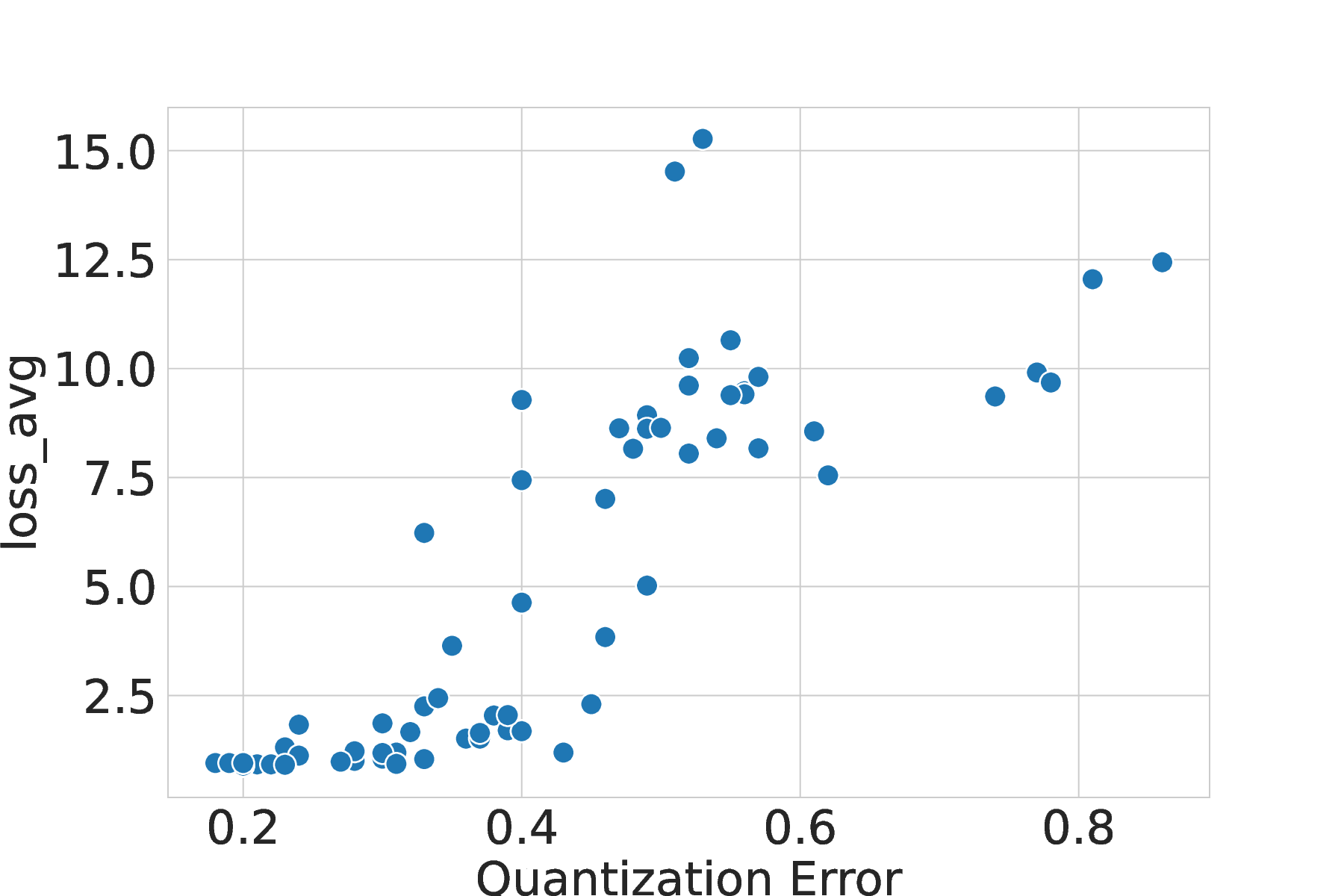}
         \caption{Average model loss vs. quantization error}
         \label{fig:loss}
     \end{subfigure}
     \caption{Relationship between quantization error and model accuracy \& loss}
     \label{fig:QE_Acc_Loss}
     \vspace{-5mm}
\end{figure}

\paragraph{Approximating Roundoff Error} We use the quantization error, QE, (measured as the MSE between f32 and dequantized weights) to approximate the impact of quantization on model accuracy for three reasons. First, prior works have leveraged QE as a proxy for quantized model accuracy - \cite{Banner2019PostT4} used quantization MSE to approximate optimal clipping value (ACIQ) and optimal bit-width for each channel. Second, we provide empirical evidence that there is a negative relationship between model accuracy and QE (see Figure~\ref{fig:QE_Acc_Loss}). Third, computing layer-wise QE instead of determining the model accuracy with respect to each layer and each possible layer bit-width has the advantage of linear time complexity. An exhaustive combinatorial search runs in exponential time \cite{Wu2018MixedPQ}.

\paragraph{Time Complexity Analysis} We analyze the algorithm's time complexity by considering its two logical components - the error calculations and the bit-width search based on them. Let L be the total number of layers, B the total number of bit-widths, and M the total of QEMs. We calculate the QE of each layer for each bit-width. Thus, the time complexity of \textit{MixQuant's} error calculations (line 4-11 in Algorithm~\ref{alg:MixQuant}) is $\mathcal{O}(L*B)$. The bit-width search (line 12-13 in Algorithm~\ref{alg:MixQuant}) compares the QE of each bit-width to the baseline int8 QE for each layer and can be performed for M number of QEMs, which takes $\mathcal(L*(B*M))$. Therefore, the overall time complexity is $\mathcal{O}(L*B) + \mathcal{O}(L(B*M)) = \mathcal{O}(L(B+B*M))$.
This is linear with respect to the number of layers. If we used model loss instead of layer QE to search for optimal bits, we would need to consider all the models generated via the combinations of B number of bit-widths over L number of layers. The time complexity would be $\mathcal{O}(B^L)$, which is exponential.

\section{Results}
We implement \textit{MixQuant} using Python and combine it with two types of quantization techniques: (1) BRECQ \cite{Li2021BRECQPT}, a state-of-the-art quantization method, and (2) vanilla asymmetric quantization \cite{Jacob2018QuantizationAT} and evaluate it on the validation set of the Imagenet ILSVRC2012 dataset. Our results demonstrate that \textit{MixQuant} can optimize the performance of existing quantization techniques. 

\paragraph{\textit{MixQuant} with BRECQ}
BRECQ is a state-of-the art quantization method that has been shown to outperform OMSE \cite{Choukroun2019LowbitQO}, AdaRound \cite{Nagel2020UpOD}, AdaQuant \cite{Hubara2020ImprovingPT}, and Bit-Split \cite{Wang2020TowardsAP}, and in Table \ref{table:results}, we demonstrate in that when \textit{MixQuant} is combined with BRECQ, we achieve better quantized accuracy than BRECQ alone. Additionally, Table \ref{table:results} also compares our results with \cite{Liu2021PosttrainingQW}, a state of the art mixed precision quantization technique, and shows that the accuracy degradation is significantly greater in \cite{Liu2021PosttrainingQW}.

\begin{table}[]
\vspace{-5mm}
\caption{Comparison of \textit{MixQuant} combined with BRECQ, BRECQ, and \cite{Liu2021PosttrainingQW}}
\label{table:results}
\tiny
\centering
\begin{tabular}{c|lr|lr|rr}
\multicolumn{1}{l|}{}        & \textbf{Bits W/A}      & \multicolumn{1}{l|}{\textbf{MixQuant + BRECQ}} & Bits W/A & \multicolumn{1}{l|}{BRECQ} & \multicolumn{1}{l}{Bits W/A} & \multicolumn{1}{l}{Liu et al. (2021)} \\ \hline
\multirow{4}{*}{ResNet-18}   & \textbf{32/32}         & \textbf{69.76}                                 & 32/32    & 71.08                      & \multicolumn{1}{l}{32/32}    & 74.24                                 \\
                             & \textbf{4, 5, 6/32}    & \textbf{70.69}                                 & 4/32     & 70.7                       & 4/8                          & 61.68                                 \\
                             & \textbf{4, 5, 6/32}    & \textbf{70.69}                                 & 3/32     & 69.81                      & 4/8                          & 61.68                                 \\
                             & \textbf{2, 5, 6/32}    & \textbf{68.93}                                 & 2/32     & 66.3                       & 4/8                          & 61.68                                 \\ \hline
\multirow{4}{*}{MobileNetV2} & \textbf{32/32}         & \textbf{71.88}                                 & 32/32    & 72.49                      & \multicolumn{1}{l}{}         & 71.78                                 \\
                             & \textbf{4, 5, 6, 7/32} & \textbf{71.92}                                 & 4/32     & 71.66                      & 8/8                          & 70.7                                  \\
                             & \textbf{4, 5, 6, 7/32} & \textbf{71.92}                                 & 3/32     & 69.5                       & 8/8                          & 70.7                                  \\
                             & \textbf{2, 5, 6/32}    & \textbf{59.53}                                 & 2/32     & 59.67                      & 8/8                          & 70.7                                 
\end{tabular}
\vspace{-5mm}
\end{table}

\paragraph{\textit{MixQuant} with Asymmetric Quantization}
In addition to BRECQ, we combine \textit{MixQuant} with asymmetric quantization and compare its quantized model accuracy with f32 and int8 baselines. Table~\ref{table:MixQuantResults} shows the set of bit-widths found via \textit{MixQuant} for various QEMs and various ResNet architectures along with model top-1 and top-5 accuracy. A user can flexibly select the quantization solution based on \textit{their} requirements with the QEM. For higher QEMs the bit-widths are lower and the model accuracy decreases, while for lower QEMs the bit-widths and quantized model accuracy are higher. Therefore, \textit{MixQuant} allows its user to flexibly select the trade-off between model accuracy and lowering the quantization bit-width. For example, the highlighted lines in Table~\ref{table:MixQuantResults} satisfy the requirement of selecting the minimum quantization bit-widths such that the model top-1 accuracy degradation is $\leq$ 3\%.

\begin{table*}[]
\vspace{-5mm}
\centering
\caption{\textit{MixQuant} with vanilla asymmetric quantization}
\label{table:MixQuantResults}
\tiny
\begin{tabular*}{\textwidth}{p{2.0cm}p{1.5cm}p{0.8cm}p{2.6cm}p{1.2cm}p{1.2cm}p{1.2cm}p{1.2cm}}
\textbf{Architecture} & \multicolumn{1}{l}{\textbf{Experiment}} & \multicolumn{1}{l}{\textbf{QEM}} & \textbf{layers\_bit\_widths} & \multicolumn{1}{l}{\textbf{Top-1 Acc}} & \multicolumn{1}{l}{\textbf{Top-5 Acc}} & \multicolumn{1}{l}{\textbf{Avg Loss}} & \multicolumn{1}{l}{\textbf{QMSE}}  \\ 
\hline
resnet18              & \multicolumn{1}{l}{baseline: f32}       & \multicolumn{1}{l}{N/A}          & all layers are float 32      & 69.76                               & 89.08                               & 1.25                                    & \multicolumn{1}{l}{N/A}            \\
resnet18              & \multicolumn{1}{l}{baseline: int8}      & \multicolumn{1}{l}{N/A}          & all layers are int 8         & 69.63                               & 89.07                               & 1.25                                    & \multicolumn{1}{l}{N/A}            \\ 
\hline
\textcolor{blue}{resnet18}              & \multirow{5}{*}{MixQuant}                & \textcolor{blue}2                                 & \textcolor{blue}{6, 7}                         & \textcolor{blue}{68.20}                               & \textcolor{blue}{88.30}                               & \textcolor{blue}{1.31}                                    & \textcolor{blue}{0.23}                               \\
resnet18              &                                          & 3                                 & 5, 6, 7                      & 63.96                               & 85.58                               & 1.51                                    & 0.36                               \\
resnet18              &                                          & 3.25                              & 5, 6                         & 64.00                               & 85.54                               & 1.51                                    & 0.37                               \\
{resnet18}              &                                          & {3.3}                               & {4, 5, 6}                     & {61.29}                               & {83.81}                               & {1.64}                                    & {0.37}                               \\
resnet18              &                                          & 3.5                               & 4, 6                         & 53.67                               & 77.78                               & 2.04                                    & 0.38                               \\ 
\hline\hline
resnet34              & \multicolumn{1}{l}{baseline: f32}       & \multicolumn{1}{l}{N/A}          & all layers are float 32      & 73.31                               & 91.42                               & 1.08                                    & \multicolumn{1}{l}{N/A}            \\
resnet34              & \multicolumn{1}{l}{baseline: int8}      & \multicolumn{1}{l}{N/A}          & all layers are int 8         & 73.24                               & 91.39                               & 1.08                                    & \multicolumn{1}{l}{N/A}            \\ 
\hline
\textcolor{blue}{resnet34}              & \multirow{4}{*}{MixQuant}                & \textcolor{blue}2                                 & \textcolor{blue}{6, 7}                         & \textcolor{blue}{72.35}                               & \textcolor{blue}{90.91}                               & \textcolor{blue}{1.12}                                    & \textcolor{blue}{0.24}                               \\
resnet34              &                                          & 3                                 & 4, 5, 6, 7                   & 61.21                               & 82.93                               & 1.70                                    & 0.39                               \\
resnet34              &                                          & 3.25                              & 4, 6                         & 61.36                               & 83.05                               & 1.68                                    & 0.40                               \\
{resnet34}              &                                          & {3.3}                               & {4, 6}                         & {61.36}                               & {83.05}                               & {1.68}                                    & {0.40}                               \\ 
\hline\hline
resnet50              & \multicolumn{1}{l}{baseline: f32}       & \multicolumn{1}{l}{N/A}          & all layers are float 32      & 76.13                               & 92.86                               & 0.96                                    & \multicolumn{1}{l}{N/A}            \\
resnet50              & \multicolumn{1}{l}{baseline: int8}      & \multicolumn{1}{l}{N/A}          & all layers are int 8         & 75.99                               & 92.81                               & 0.97                                    & \multicolumn{1}{l}{N/A}            \\ 
\hline
\textcolor{blue}{resnet50}              & \multirow{3}{*}{MixQuant}                & \textcolor{blue}2                                 & \textcolor{blue}{6, 7}                         & \textcolor{blue}{75.18}                               & \textcolor{blue}{92.52}                               & \textcolor{blue}{1.00}                                    & \textcolor{blue}{0.28}                               \\
{resnet50}              &                                          & {3}                                 & {4, 5, 6}                      & {70.58}                               & {90.04}                               & {1.19}                                    & {0.43}                               \\
resnet50              &                                          & 3.25                              & 4, 5, 6                      & 50.13                               & 74.29                               & 2.30                                    & 0.45                               \\ 
\hline\hline
resnet101             & \multicolumn{1}{l}{baseline: f32}       & \multicolumn{1}{l}{N/A}          & all layers are float 32      & 77.37                               & 93.55                               & 0.91                                    & \multicolumn{1}{l}{N/A}            \\
resnet101             & \multicolumn{1}{l}{baseline: int8}      & \multicolumn{1}{l}{N/A}          & all layers are int 8         & 77.21                               & 93.51                               & 0.92                                    & \multicolumn{1}{l}{N/A}            \\ 
\hline
\textcolor{blue}{resnet101}             & \multirow{5}{*}{MixQuant}                & \textcolor{blue}{1.3}                               & \textcolor{blue}{5, 6, 7, 8}                   & \textcolor{blue}{76.96}                               & \textcolor{blue}{93.42}                               & \textcolor{blue}{0.92}                                    & \textcolor{blue}{0.22}                               \\
resnet101             &                                          & 1.5                               & 2, 5, 6, 7, 8                & 59.23                               & 81.74                               & 1.83                                    & 0.24                               \\
resnet101             &                                          & 1.7                               & 2, 3, 4, 5, 6, 7, 8          & 58.86                               & 81.05                               & 1.86                                    & 0.30                               \\
resnet101             &                                          & 1.8                               & 2, 3, 5, 6, 7                & 52.32                               & 75.61                               & 2.25                                    & 0.33                               \\
resnet101             &                                          & 1.9                               & 2, 4, 5, 6, 7                & 49.36                               & 72.63                               & 2.44                                    & 0.34                               \\ 
\hline\hline
resnet152             & \multicolumn{1}{l}{baseline: f32}       & \multicolumn{1}{l}{N/A}          & all layers are float 32      & 78.31                               & 94.05                               & 0.88                                    & \multicolumn{1}{l}{N/A}            \\
resnet152             & \multicolumn{1}{l}{baseline: int8}      & \multicolumn{1}{l}{N/A}          & all layers are int 8         & 78.31                               & 94.02                               & 0.88                                    & \multicolumn{1}{l}{N/A}            \\ 
\hline
resnet152             & \multirow{6}{*}{MixQuant}                & 1.1                               & 7, 8                         & 78.20                               & 94.01                               & 0.88                                    & 0.20                               \\
resnet152             &                                          & 1.3                               & 6, 7, 8                      & 78.15                               & 94.01                               & 0.89                                    & 0.20                               \\
\textcolor{blue}{resnet152}             &                                          & \textcolor{blue}{1.5}                               & \textcolor{blue}{5, 6, 7, 8}                   & \textcolor{blue}{77.58}                               & \textcolor{blue}{93.76}                               & \textcolor{blue}{0.91}                                    & \textcolor{blue}{0.23}                               \\
resnet152             &                                          & 1.7                               & 3, 5, 6, 7, 8                & 70.68                               & 90.11                               & 1.22                                    & 0.28                               \\
{resnet152}             &                                          & {1.8}                               & {2, 5, 6, 7}                   & {71.48}                               & {90.16}                               & {1.19}                                    & {0.31}                               \\
resnet152             &                                          & 1.9                               & 2, 4, 5, 6, 7                & 62.99                               & 85.01                               & 1.66                                    & 0.32                               \\ 
\hline\hline
resnext50\_32x4d      & \multicolumn{1}{l}{baseline: f32}       & \multicolumn{1}{l}{N/A}          & all layers are float 32      & 77.62                               & 93.70                               & 0.94                                    & \multicolumn{1}{l}{N/A}            \\
resnext50\_32x4d      & \multicolumn{1}{l}{baseline: int8}      & \multicolumn{1}{l}{N/A}          & all layers are int 8         & 77.40                               & 93.63                               & 0.95                                    & \multicolumn{1}{l}{N/A}            \\ 
\hline
resnext50\_32x4d      & \multirow{6}{*}{MixQuant}                & 1.3                               & 7, 8                         & 77.43                               & 93.52                               & 0.95                                    & 0.19                               \\
resnext50\_32x4d      &                                          & 1.5                               & 6, 7, 8                      & 77.21                               & 93.51                               & 0.95                                    & 0.20                               \\
resnext50\_32x4d      &                                          & 1.7                               & 5, 6, 7, 8                   & 76.93                               & 93.29                               & 0.98                                    & 0.27                               \\
resnext50\_32x4d      &                                          & 1.8                               & 5, 6, 7                      & 75.43                               & 92.60                               & 1.05                                    & 0.30                               \\
\textcolor{blue}{resnext50\_32x4d}      &                                          & \textcolor{blue}{1.9}                               & \textcolor{blue}{5, 6, 7}                      & \textcolor{blue}{75.43}                               & \textcolor{blue}{92.60}                               & \textcolor{blue}{1.05}                                    & \textcolor{blue}{0.30}                               \\
{resnext50\_32x4d}      &                                          & {2}                                 & {4, 5, 6, 7}                   & {72.60}                               & {90.79}                               & {1.18}                                    & {0.30}                               \\ 
\hline\hline
resnext101\_32x8d     & \multicolumn{1}{l}{baseline: f32}       & \multicolumn{1}{l}{N/A}          & all layers are float 32      & 79.31                               & 94.53                               & 0.93                                    & \multicolumn{1}{l}{N/A}            \\
resnext101\_32x8d     & \multicolumn{1}{l}{baseline: int8}      & \multicolumn{1}{l}{N/A}          & all layers are int 8         & 79.11                               & 94.51                               & 0.93                                    & \multicolumn{1}{l}{N/A}            \\ 
\hline
resnext101\_32x8d     & \multirow{5}{*}{MixQuant}                & 1.1                               & 7, 8                         & 79.12                               & 94.51                               & 0.93                                    & 0.31                               \\
\textcolor{blue}{resnext101\_32x8d}     &                                          & \textcolor{blue}{1.3}                               & \textcolor{blue}{4, 6, 7, 8}                   & \textcolor{blue}{76.61}                               & \textcolor{blue}{93.26}                               & \textcolor{blue}{1.04}                                    & \textcolor{blue}{0.33}                               \\
resnext101\_32x8d     &                                          & 1.5                               & 2, 4, 5, 6, 7, 8             & 59.91                               & 81.46                               & 2.05                                    & 0.39                               \\
resnext101\_32x8d     &                                          & 1.7                               & 2, 4, 5, 6, 7, 8             & 37.65                               & 59.52                               & 3.84                                    & 0.46                               \\
resnext101\_32x8d     &                                          & 1.8                               & 2, 3, 4, 5, 6, 7             & 26.14                               & 45.57                               & 5.02                                    & 0.49             \\                 
\hline
\end{tabular*}
\vspace{-5mm}
\end{table*}

\paragraph{Runtime Analysis}
\label{subsec:runtime}
Table~\ref{table:MixQuantRuntime} reports the runtime in seconds of \textit{MixQuant} for various ResNet architectures, where \textit{MixQuant} considers the bit-widths of 8, 7, 6, 5, 4, 3, and 2, and one or ten different QEMs. It can be observed that the runtime grows with the number of layers since higher number of layers imply a larger search space. For one QEM, the \textit{MixQuant} search takes between 0.1 and 0.5 seconds. If it is combined with asymmetric per-layer quantization using the optimal bit-widths returned by the search, it takes between 1.0 and 3.2 seconds. If the number of QEMs is increased from one to ten the \textit{MixQuant} search takes between 0.9 and 5.5 seconds, which represents a linear increase in runtime.

\begin{table}[]
\vspace{-8mm}
    \tiny
    \caption{Runtime of \textit{MixQuant} for (a) 1 QEM and (b) 10 QEMs}
    \vspace{-3mm}
    \label{table:MixQuantRuntime}
    \begin{subtable}[b]{.45\linewidth}
        \caption{}
            
            \begin{tabular}{p{2.0cm}p{0.7cm}p{0.8cm}}
            \textbf{Architecture} & \multicolumn{1}{l}{\textbf{search}} & \multicolumn{1}{l}{\textbf{search + quantization}}  \\ 
            \hline
            resnet18              & 0.1 s                                  & 1 s                                             \\
            resnet34              & 0.2 s                                 & 1.1 s                                                \\
            resnet50              & 0.2 s                                 & 1.3 s                                                \\
            resnet101             & 0.4 s                                 & 1.7 s                                                \\
            resnet152             & 0.5 s                                 & 2  s                                                 \\
            resnext50\_32x4d      & 0.2 s                                 & 1.4  s                                               \\
            resnext101\_32x8d     & 0.5 s                                 & 3.2 s                                                \\
            \hline
            \end{tabular}
    \end{subtable}
    \hspace{2.5em}
    \begin{subtable}[b]{.45\linewidth}
        \caption{}

            \begin{tabular}{p{2.0cm}p{0.7cm}p{0.8cm}}
            \textbf{Architecture} & \multicolumn{1}{l}{\textbf{search}} & \multicolumn{1}{l}{\textbf{search+quantization}}  \\ 
            \hline
            resnet18              & 0.9 s                                & 1.8 s                                               \\
            resnet34              & 1.5 s                             & 2.5 s                                              \\
            resnet50              & 2 s                                  & 3.1 s                                               \\
            resnet101             & 3.6 s                                 & 4.9 s                                               \\
            resnet152             & 5.2 s                                 & 6.8 s                                              \\
            resnext50\_32x4d      & 2 s                                  & 3.2 s                                              \\
            resnext101\_32x8d     & 5.5 s                                & 8.2 s                                            \\
            \hline
            \end{tabular}
    \end{subtable} 
    \vspace{-5mm}
\end{table}
\vspace{-5mm}

\section{Quantization Sensitivity of Weights Ablation Studies}
To demonstrate that quantizing DNN weights warrants a search for optimal bit-widths as opposed to uniform precision quantization, we perform two ablation studies to show that different weight layers have different sensitivity to quantization based on their type and position.

\paragraph{Weights Quantization Sensitivity by Layer Type}
First, we investigate if different layer types have different sensitivity to quantization. We consider four layer types in the ResNet architecture: (1) first conv layer, (2) conv layers with a 3x3 kernel, (3) conv layers with a 1x1 kernel, and (4) final fully connected layer. For each type of layer, we perform  asymmetric quantization and vary its bit-width while keeping the bit-width of all other layer types constant at int8. We calculate the model accuracy, loss, and quantization error for the following quantization bit-widths: 8, 7, 6, 5, 4, 3, and 2. 

\begin{figure*}[]
\vspace{-5mm}
     \centering
     \captionsetup[subfigure]{justification=centering}
     \begin{subfigure}[b]{0.3\textwidth}
         \centering
         \includegraphics[width=\linewidth]{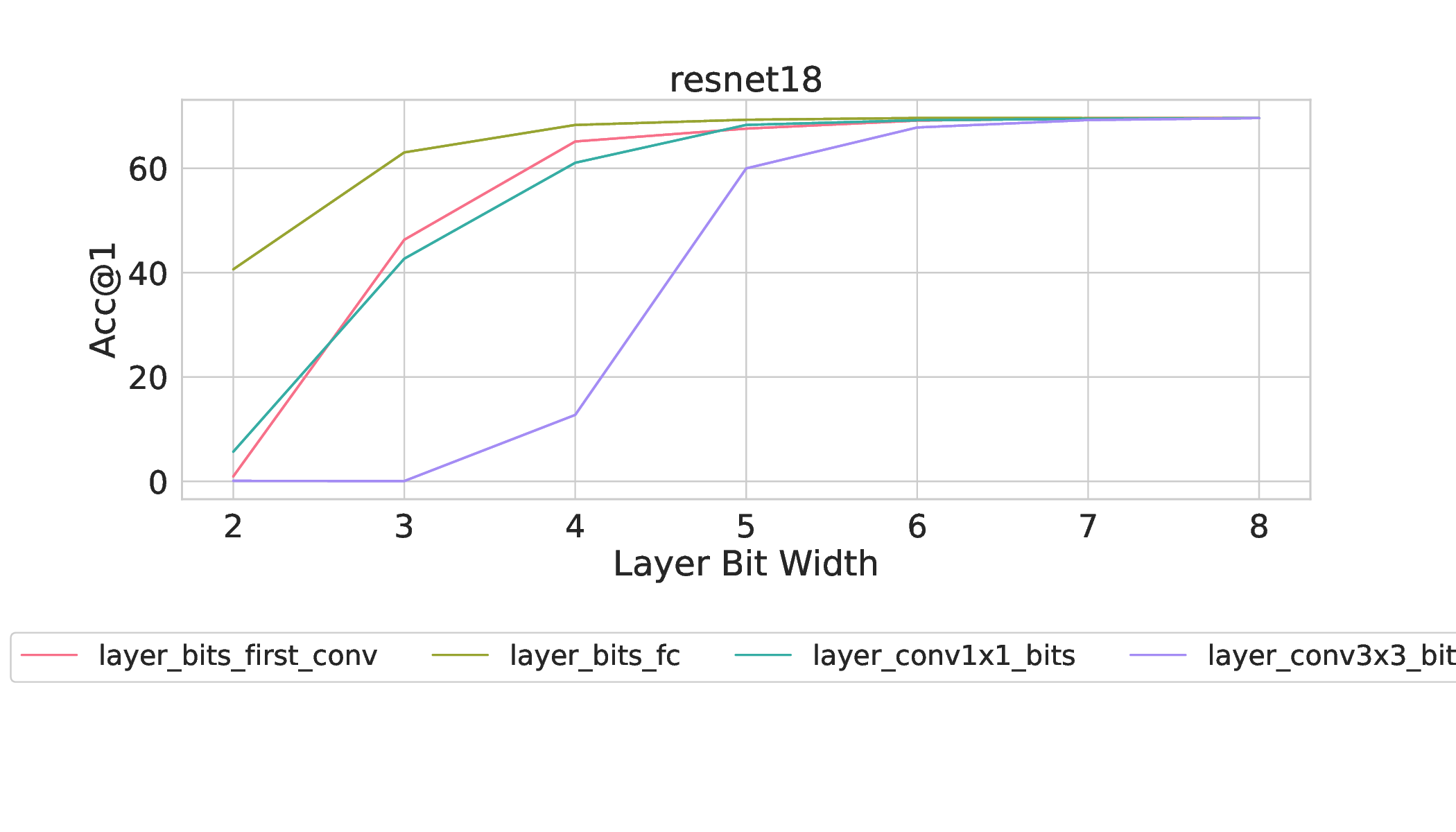}
         \vspace*{-10mm}
         \caption{ResNet18}
     \end{subfigure}
     \begin{subfigure}[b]{0.3\textwidth}
         \centering
         \includegraphics[width=\linewidth]{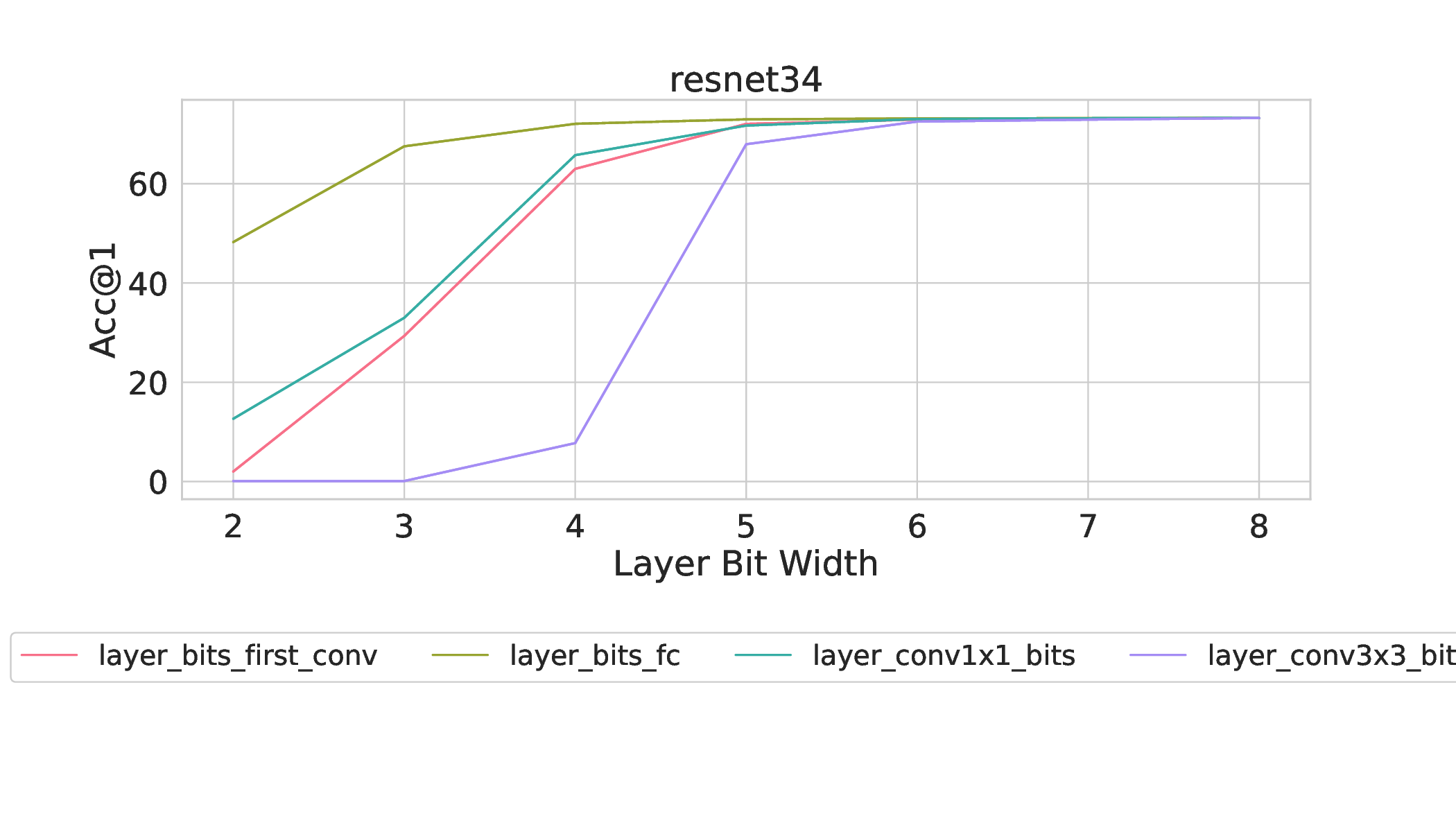}
         \vspace*{-10mm}
         \caption{ResNet34}
     \end{subfigure}
     \begin{subfigure}[b]{0.3\textwidth}
         \centering
         \includegraphics[width=\linewidth]{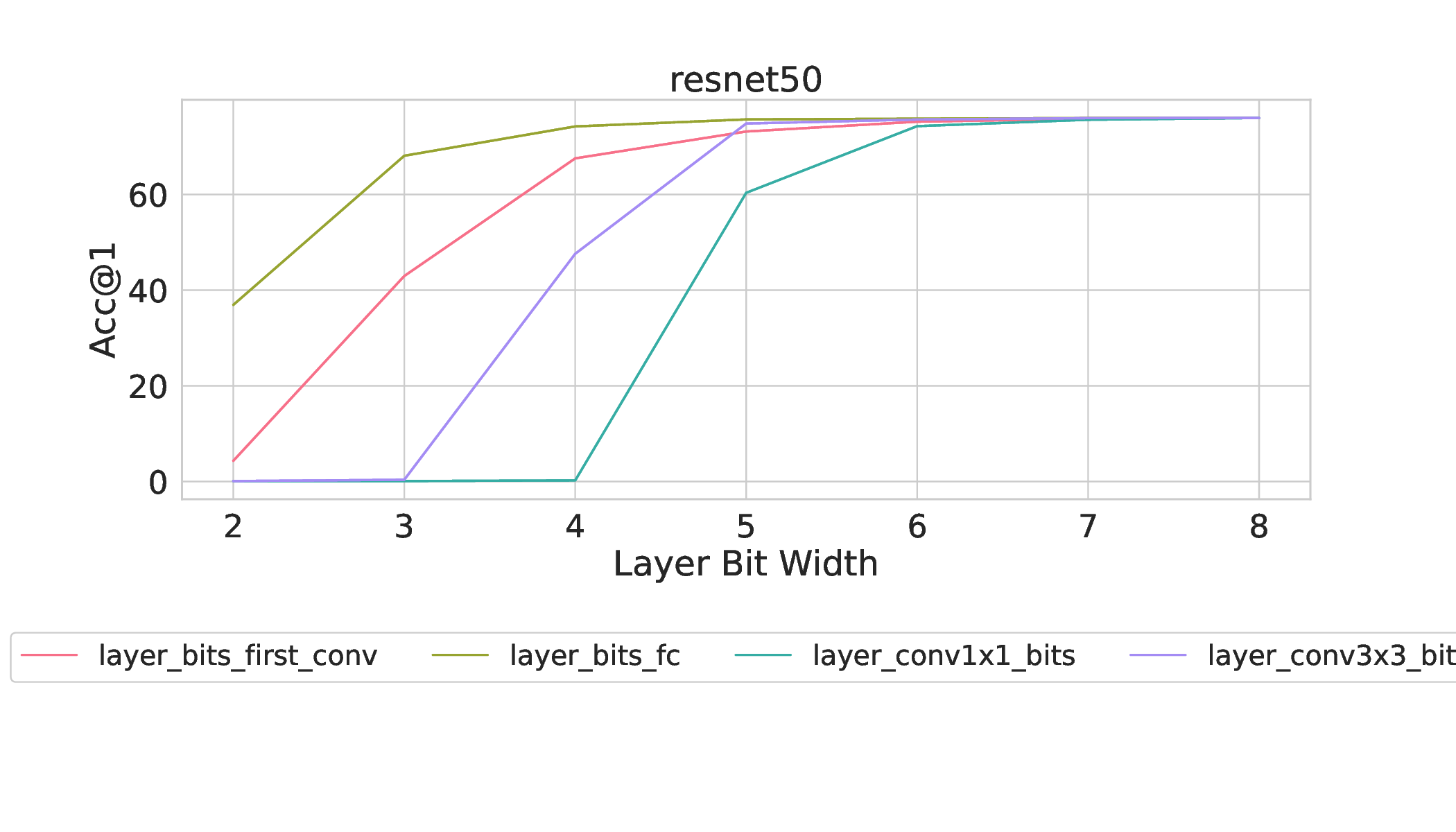}
         \vspace*{-10mm}
         \caption{ResNet50}
     \end{subfigure}
     
     \begin{subfigure}[b]{0.45\textwidth}
         \centering
         \includegraphics[width=\linewidth]{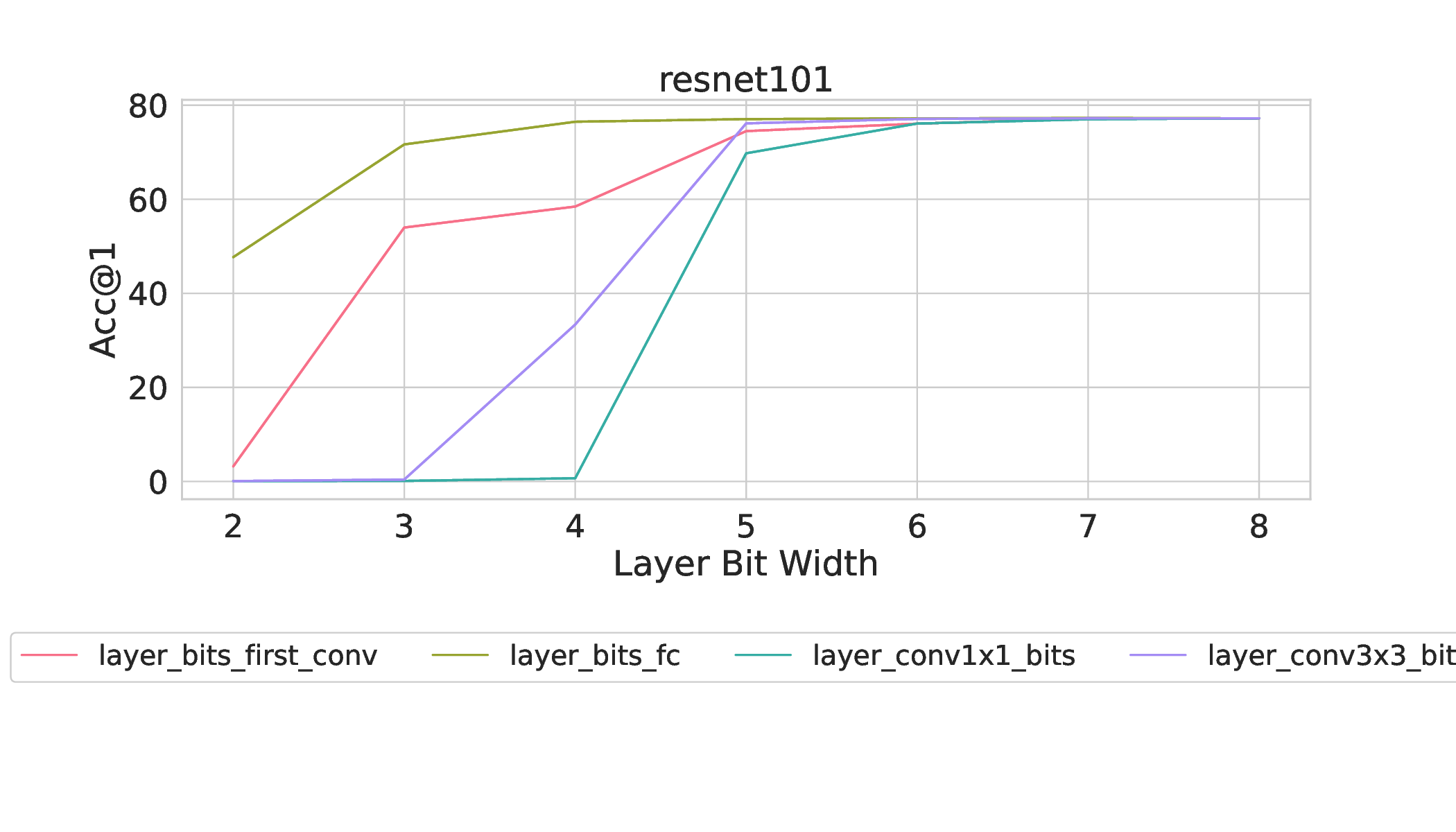}
         \vspace*{-10mm}
         \caption{ResNet101}
     \end{subfigure}
     \begin{subfigure}[b]{0.45\textwidth}
         \centering
         \includegraphics[width=\linewidth]{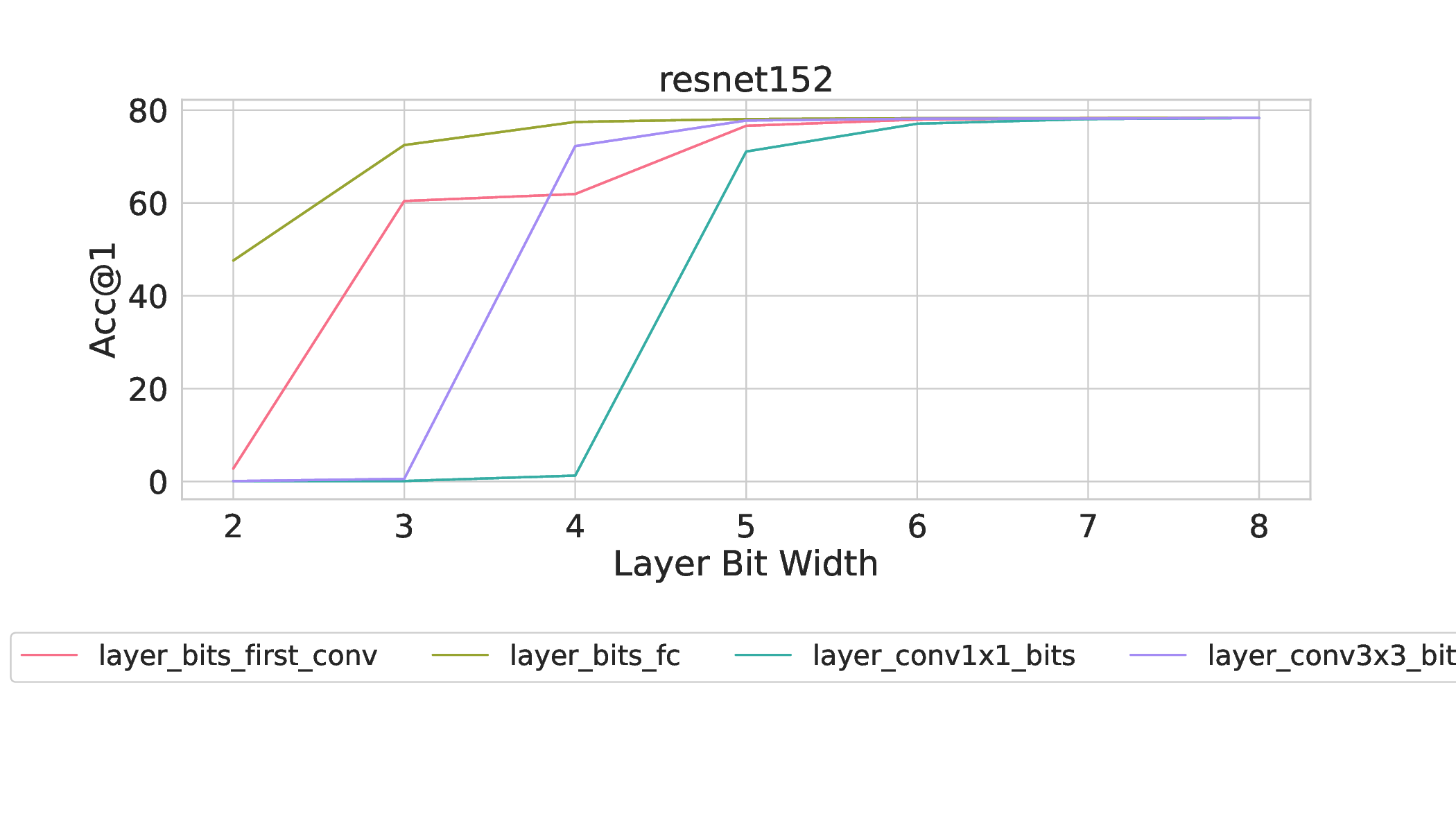}
         \vspace*{-10mm}
         \caption{ResNet152}
     \end{subfigure}
     
     \begin{subfigure}[b]{0.45\textwidth}
         \centering
         \includegraphics[width=\linewidth]{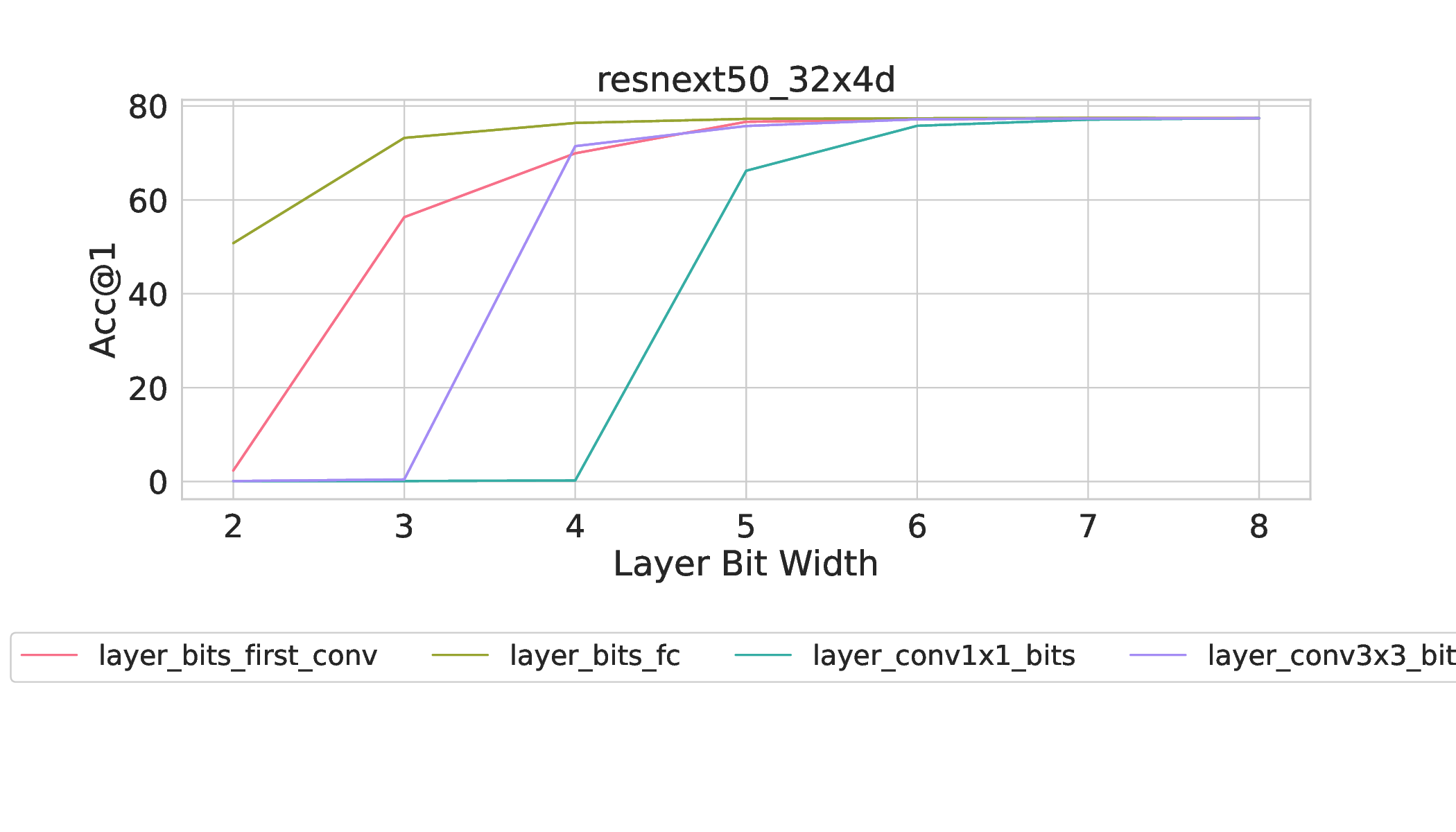}
         \vspace*{-10mm}
         \caption{ResNeXt50\_32x4d}
     \end{subfigure}
         \begin{subfigure}[b]{0.45\textwidth}
         \centering
         \includegraphics[width=\linewidth]{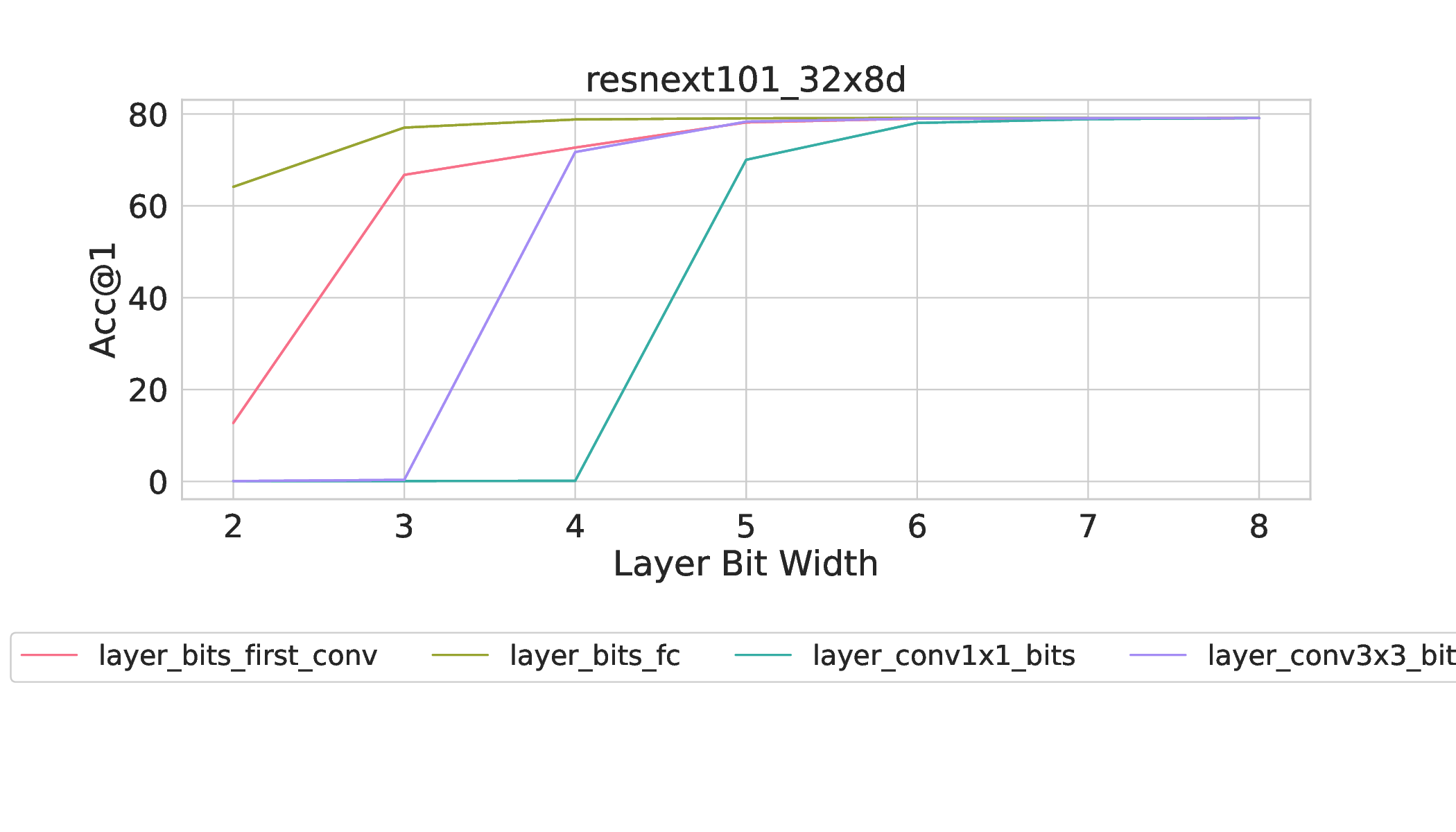}
         \vspace*{-10mm}
         \caption{ResNeXt101\_32x8d}
     \end{subfigure}
     
    \caption{Sensitivity of different layer types to quantization}
    \label{fig:LayerTypeSensitivity_Acc_vs_Architecture}
    \vspace{-5mm}
\end{figure*}

In Figure~\ref{fig:LayerTypeSensitivity_Acc_vs_Architecture}, we show the impact of varying the bit-width of one layer type at a time on the model top-1 accuracy. Lowering the quantization bit-width of conv layers with a 3x3 kernel has the most adverse impact on top-1 accuracy in shallower ResNet architectures, while in deeper ones it is the conv layers with a 1x1 kernel followed by conv layers with a 3x3 kernel that impacts model accuracy the most. The first conv layer and conv layers with a 1x1 kernel have approximately the same sensitivity to varying bit-width in the shallower architectures. Finally, the quantization bit-width of the final fully connected layer has the smallest impact on model accuracy for all ResNet architectures. In general, starting at 5 bits the model accuracy begins to degrade; however, the deeper architectures are less sensitive to decreasing bit-width. 
While the reason that the conv layers with a 3x3 kernel and 1x1 kernel are the most sensitive is the fact that those layer types account for the highest number of layers in ResNet, we can still conclude that different layer types have different sensitivity to quantization bit-width measured as the impact on the overall model quality. Therefore, different layer types will benefit from different quantization bit-widths, which motivates \textit{MixQuant}. Similar results can also be found by measuring layer type sensitivity using the model average loss and quantization mean squared error.

\paragraph{Weights Quantization Sensitivity by Layer Position}
In addition to the layer type, we investigate if the position of a layer has an impact on quantization sensitivity of weights. We measure the {\it relative quantization error} (RQE) of individual layers for the following bit-widths: 8, 7, 6, 5, 4, 3, 2, and define the RQE as $RQE = avg ((\vec{f32 \,w} - \vec{dequantized \,w}) / \vec{f32 \,w})$, where $\vec{w}$ is the weights vector and the $avg$ operation returns a scalar that represents the mean of all elements in a vector.

\begin{table}
\vspace{-5mm}
\tiny
\centering
\caption{The most sensitive layer positions in a DNN measured as a relative quantization error with respect to varying quantization bit-width}
\label{table:Sensitivity_layer_position}
\begin{tabular}{l|r|r|r|r|r|r|r}
\textbf{Architecture} & \multicolumn{1}{l|}{\textbf{int8}} & \multicolumn{1}{l|}{\textbf{int7}} & \multicolumn{1}{l|}{\textbf{int6}} & \multicolumn{1}{l|}{\textbf{int5}} & \multicolumn{1}{l|}{\textbf{int4}} & \multicolumn{1}{l|}{\textbf{int3}} & \multicolumn{1}{l}{\textbf{int2}} \\ \hline
resnet18              & 1                                  & 1                                  & 1                                  & 17                                 & 17                                 & 17                                 & 16                                \\
resnet34              & 1                                  & 1                                  & 20                                 & 20                                 & 20                                 & 33                                 & 35                                \\
resnet50              & 46                                 & 46                                 & 46                                 & 46                                 & 46                                 & 47                                 & 44                                \\
resnet101             & 6                                  & 6                                  & 6                                  & 6                                  & 97                                 & 97                                 & 99                                \\
resnet152             & 1                                  & 45                                 & 45                                 & 148                                & 148                                & 148                                & 152                               \\
resnet50\_32x4d       & 1                                  & 1                                  & 1                                  & 52                                 & 45                                 & 45                                 & 45                                \\
resnet101\_32x8d      & 1                                  & 1                                  & 49                                 & 49                                 & 49                                 & 96                                 & 96                                \\ \hline
\end{tabular}
\vspace{-5mm}
\end{table}

Table~\ref{table:Sensitivity_layer_position} identifies the most sensitive layers across various bit-widths and architectures, where layers are indexed from 0 through n, and n equals is the total number of layers minus one. For example, for int8, it is the 1st layer in resnet18 that has the highest relative QE compared to all other resnet18 layers while for resnet50 it is the 46th layer. We can see that the quantization bit-width has a significant impact on the position of the most sensitive layer with the exception of ResNet50. While ResNet50's most sensitive layer is located towards the end of the network for all bit-widths, other architectures's most sensitive layer position varies based on the bit-width. For higher bit-widths 8, 7, and 6 it is located at the beginning while for lower bit-widths 2, 3, and 4 it is at the end. The most sensitive layers of ResNet34 and ResNeXt101\_32x8d at bit-widths 4, 5, and 6 are in the middle of the network. Based on these experiments, we can conclude that different layer positions have different sensitivity to varying bit-width.

\section{Conclusion}
In this paper we propose \textit{MixQuant}, a search algorithm that finds the optimal quantization bit-width for each layer weight and can be combined with any quantization method as a form of pre-processing optimization. We show that combining \textit{MixQuant} with BRECQ \cite{Li2021BRECQPT}, a state-of-the-art quantization method, yields better quantized model accuracy than BRECQ alone. Additionally, we combine BREQ with asymmetric quantization \cite{Jacob2018QuantizationAT} to show that \textit{MixQuant} has the potential to optimize the performance of any quantization technique. Our code is open-sourced and available at: \url{https://anonymous.4open.science/r/qantizedImagenet-43C5}.

%
%
%
%

\end{document}